\documentclass[letterpaper, 10 pt, conference]{IEEEtran}

\usepackage{comment}
\usepackage{siunitx}
\usepackage{relsize}
\usepackage{ifthen}
\usepackage[colorinlistoftodos]{todonotes}

\usepackage{graphics} %
\usepackage{rotating}
\usepackage{color}
\usepackage{enumerate}
\usepackage[T1]{fontenc}
\usepackage{psfrag}
\usepackage{epsfig} %
\usepackage{booktabs}
\usepackage{graphicx,url}
\usepackage{multirow}
\usepackage{array}
\usepackage{latexsym}
\usepackage{amsfonts}
\usepackage{amsmath}
\usepackage{amssymb}
\usepackage{xstring}
\usepackage{multirow}
\usepackage{xcolor}
\usepackage{prettyref}
\usepackage{flexisym}
\usepackage{bigdelim}
\usepackage{breqn} %
\usepackage{listings}
\let\labelindent\relax
\usepackage{enumitem}
\usepackage{xspace}
\usepackage{bm}
\graphicspath{{./figures/}}
\usepackage{tikz}
\usetikzlibrary{matrix,calc}

\usepackage{mdwlist}

\let\itemize\undefined
\makecompactlist{itemize}{stditemize}

\usepackage{etex}
\usepackage{algpseudocode}
\usepackage{algorithm}
\usepackage{subcaption}
\usepackage{amsmath}

\newcommand{\refining}{annealing\xspace}
\DeclareCaptionLabelSeparator{periodspace}{.\quad}
\IEEEoverridecommandlockouts  

\usepackage[letterpaper, left=0.75in, right=0.75in, bottom=0.75in, top=0.75in]{geometry}
\usepackage{multicol}
\usepackage[bookmarks=true]{hyperref}
\usepackage{amsthm}
\newtheoremstyle{mystyle}%
  {}%
  {}%
  {\itshape}%
  {}%
  {\bfseries}%
  {.}%
  { }%
  {\thmname{#1}\thmnumber{ #2}\thmnote{ (#3)}}%
\theoremstyle{mystyle}
\newrefformat{prob}{Problem\,\ref{#1}}
\newrefformat{def}{Definition\,\ref{#1}}
\newrefformat{sec}{Section\,\ref{#1}}
\newrefformat{sub}{Section\,\ref{#1}}
\newrefformat{prop}{Proposition\,\ref{#1}}
\newrefformat{app}{Appendix\,\ref{#1}}
\newrefformat{alg}{Algorithm\,\ref{#1}}
\newrefformat{cor}{Corollary\,\ref{#1}}
\newrefformat{thm}{Theorem\,\ref{#1}}
\newrefformat{lem}{Lemma\,\ref{#1}}
\newrefformat{fig}{Fig.\,\ref{#1}}
\newrefformat{tab}{Table\,\ref{#1}}

\newcommand{\bdmath}{\begin{dmath}}
\newcommand{\edmath}{\end{dmath}}
\newcommand{\beq}{\begin{equation}}
\newcommand{\eeq}{\end{equation}}
\newcommand{\bdm}{\begin{displaymath}}
\newcommand{\edm}{\end{displaymath}}
\newcommand{\bea}{\begin{eqnarray}}
\newcommand{\eea}{\end{eqnarray}}
\newcommand{\beal}{\beq \begin{array}{ll}}
\newcommand{\eeal}{\end{array} \eeq}
\newcommand{\beas}{\begin{eqnarray*}}
\newcommand{\eeas}{\end{eqnarray*}}
\newcommand{\ba}{\begin{array}}
\newcommand{\ea}{\end{array}}
\newcommand{\bit}{\begin{itemize}}
\newcommand{\eit}{\end{itemize}}
\newcommand{\ben}{\begin{enumerate}}
\newcommand{\een}{\end{enumerate}}

\newcommand{\calC}{{\cal C}}

\newcommand{\calI}{{\cal I}}

\newcommand{\calM}{{\cal M}}

\newcommand{\calO}{{\cal O}}

\newcommand{\setal}{~\emph{et~al.}\xspace}
\newcommand{\eg}{\emph{e.g.,}\xspace}

\newcommand{\myParagraph}[1]{{\bf #1.}\xspace}
\newcommand{\M}[1]{{\bm #1}} %
\renewcommand{\boldsymbol}[1]{{\bm #1}}

\newcommand{\hide}[1]{}

\newcommand{\hiddenText}{{\color{gray} hidden text.}}
\newcommand{\hideWithText}[1]{\hiddenText}

\newcommand{\prob}[1]{{\mathbb P}\left(#1\right)}

\newcommand{\diag}[1]{\mathrm{diag}\left(#1\right)}

\newcommand{\eye}{{\mathbf I}}

\newcommand{\Real}[1]{ { {\mathbb R}^{#1} } }

\newcommand{\at}[1]{^{(#1)}}

\newcommand{\SEthree}{\ensuremath{\mathrm{SE}(3)}\xspace}

\newcommand{\intexpmap}[1]{\mathrm{Exp}\left(#1\right)}

\newcommand{\expmap}[1]{\intexpmap{#1}}

\newcommand{\MX}{\M{X}}

\newcommand{\vc}{\boldsymbol{c}}

\newcommand{\vp}{\boldsymbol{p}}

\newcommand{\vr}{\boldsymbol{r}}

\newcommand{\vdelta}{\boldsymbol{\delta}}

\newcommand{\blue}[1]{{\color{blue}#1}}

\newcommand{\linkToPdf}[1]{\href{#1}{\blue{(pdf)}}}
\newcommand{\linkToPpt}[1]{\href{#1}{\blue{(ppt)}}}
\newcommand{\linkToCode}[1]{\href{#1}{\blue{(code)}}}
\newcommand{\linkToWeb}[1]{\href{#1}{\blue{(web)}}}
\newcommand{\linkToVideo}[1]{\href{#1}{\blue{(video)}}}
\newcommand{\linkToMedia}[1]{\href{#1}{\blue{(media)}}}
\newcommand{\award}[1]{\xspace} %

\usepackage{float}
\usepackage{hyperref}
\usepackage{graphicx}
\usepackage{epstopdf}
\usepackage{epsfig}
\usepackage{amsmath}
\usepackage[capitalise]{cleveref}

\newcommand{\name}{Loc-NeRF\xspace}
\newcommand{\nerf}{NeRF\xspace}
\newcommand{\nerfNav}{NeRF-Navigation\xspace}
\newcommand{\nerfs}{NeRFs\xspace}
\newcommand{\image}{\calI}
\newcommand{\odom}{\calO}
\newcommand{\map}{\calM}
\renewcommand{\at}[1]{^{#1}}
\newcommand{\alpharef}{\alpha_{{\small \text{refine}}}}
\newcommand{\alphasuperref}{\alpha_{{\small \text{super-refine}}}}
\newcommand{\sigmaRinit}{\sigma_{R,{\text{\footnotesize  init}}}}
\newcommand{\sigmatinit}{\sigma_{t,{\text{\footnotesize  init}}}}
\newcommand{\nreduced}{n_{\text{\footnotesize  reduced}}}
\usepackage{xr}

\title{\vspace{0.25in}\LARGE \bf{\name: Monte Carlo Localization \\ using  Neural Radiance Fields}}

\author{Dominic Maggio,
\thanks{D.\,Maggio is with the Laboratory for 
Information \& Decision Systems, Massachusetts Institute of Technology, Cambridge, MA, USA, 
and is a Draper Scholar with the Perception and Embedded ML Group,
Draper, Cambridge, MA, USA, 
\sf{drmaggio@mit.edu}
}Marcus Abate, Jingnan Shi,
\thanks{M.\,Abate, J.\,Shi, and L.\,Carlone are with the Laboratory for 
Information \& Decision Systems, Massachusetts Institute of Technology, Cambridge, MA, USA, 
\{\sf{mabate, jnshi, lcarlone}\}@mit.edu
}Courtney Mario,
\thanks{C.\,Mario is with the Draper Perception and Embedded ML Group,
Draper, Cambridge, MA, USA,  
\sf{cmario@draper.com}
}Luca Carlone
\thanks{This work was partially funded by the NASA Flight Opportunities under grant Nos 80NSSC21K0348.}
}

\begin{document}

\maketitle

\begin{abstract}
We present \name, a real-time vision-based robot localization approach that combines 
 Monte Carlo localization and Neural Radiance Fields (NeRF).
Our system uses a pre-trained NeRF model as the map of an environment and can localize itself in real-time using an RGB camera as the only exteroceptive sensor onboard the robot. 
While neural radiance fields have seen significant applications for visual rendering in computer vision and  graphics, they have found limited use in robotics.
Existing approaches for NeRF-based localization require both a good initial pose guess and significant computation, making them impractical for real-time robotics applications. 
By using Monte Carlo localization as a workhorse to estimate poses using a NeRF map model, \name is able to perform localization faster than the  
state of the art and without relying on an initial pose estimate.
In addition to testing on synthetic data, we also run our system using real data collected by a Clearpath
Jackal UGV and demonstrate for the first time the ability to perform real-time 
global localization with neural radiance fields. We make our code publicly available at \url{https://github.com/MIT-SPARK/Loc-NeRF}.
\end{abstract}

\section{Introduction}
\label{sec:intro}

Vision-based localization is a foundational problem in robotics and computer vision, with applications ranging from self-driving vehicles~\cite{Wang19pami-apolloscape} to robot manipulation~\cite{Manuelli19-kpam}.
Classical approaches for camera pose estimation typically address the task
by adopting a multi-stage paradigm,
where keypoints are first detected and matched between each frame and the map (where the latter is stored as a collection of images with the corresponding keypoints and descriptors),
and six degree-of-freedom (DoF) poses are estimated using Perspective-n-Point (PnP) algorithms~\cite{Manuelli19-kpam, Lepetit09-epnp, Ke20-gsnet}.
However, such methods are sensitive to the quality of the keypoint matching and require storing a database of images  as the map representation.

Orthogonal to the camera pose estimation literature,
advances in deep learning have led to a plethora of works investigating
implicit shape and scene representations~\cite{Deng22ral-icaps, Zhu22cvpr-niceslam, Park19cvpr-deepSDF, Yen20iros-inerf, Sucar21iccv-iMAP}.
In particular, Neural Radiance Fields (\nerf) have gained significant popularity,
as they can encode both 3D geometry and appearance of an environment~\cite{Mildenhall20arxiv-nerf}.
\nerfs are fully-connected neural networks trained using a collection of monocular images to approximate functions taking 3D positions as inputs and returning RGB values and view density (the so called ``radiance'') as output.
 \nerf can then be used in conjunction with ray tracing algorithms to synthesize novel views~\cite{Mildenhall20arxiv-nerf}.
\nerf has even been extended to address challenging rendering problems involving non-Lambertian surfaces, variable lighting
conditions~\cite{Martinbrualla20cvpr-nerfw}, and motion blur~\cite{Ma21arXiv-deblurNerf}.

\begin{figure}
    \includegraphics[width=0.47\textwidth]{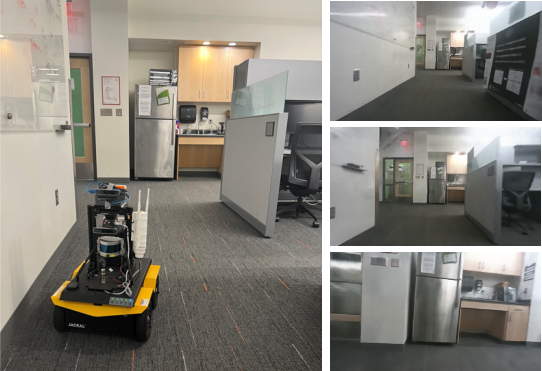}
    \caption{Real-time experiments with \name using a Clearpath
    Jackal UGV (left) equipped with a Realsense d455 camera. Examples of \nerf renderings near the beginning, middle, and end of the experiment (right).\label{fig:jackal} \vspace{-3mm}}
\end{figure}

If we view \nerf as a function that encodes spatial and radiance information,
a natural question that arises is: \emph{can we leverage advances in \nerf to solve localization tasks for robotics?}
 The existing literature on \nerf-based localization is sparse.
Yen-Chen\setal~\cite{Yen20iros-inerf} propose i\nerf, the first method to demonstrate pose estimation by ``inverting'' a \nerf; i\nerf estimates the camera pose by performing local optimization of a loss function quantifying the per-pixel  mismatch between the map and a given camera image.
Adamkiewicz\setal~\cite{Adamkiewicz21arxiv-nerfNavigation} propose \nerfNav, which demonstrates the possibility of using \nerf as a map representation across the autonomy stack, from state estimation to planning.

\myParagraph{Contributions} Following the same research thrust as i\nerf and \nerfNav,
we present \emph{\name}, a 6DoF pose estimation pipeline that uses a (particle-filter-based) Monte Carlo localization~\cite{Dellaert99}  approach as a novel way to extract poses from a \nerf.
More in detail, we design a vision-based particle-filter localization pipeline, that
(i) uses \nerf as a map model in the update step of the filter, 
and (ii) uses visual-inertial odometry or the robot dynamics for highly accurate motion estimation 
in the prediction step of the filter.
The proposed particle-filter approach allows
pose estimation with poor or no initial guess, while allowing us to adjust the computational effort by 
modifying the number of particles.
We present extensive experiments showing that \name can:
(i) estimate the pose of a single image without relying on an accurate initial guess, 
(ii) perform global localization, and
(iii) achieve real-time tracking with real-world data (\cref{fig:jackal}).

The rest of the paper is organized as follows. \cref{sec:related_works} discusses related work. 
\cref{sec:background} provides a high level overview of \nerf.
\cref{sec:contribution}  presents the structure of \name.
\cref{sec:experiments} evaluates \name on three types of experiments: 
benchmarking with i\nerf on pose estimation from a single image, 
benchmarking with \nerfNav on simulated drone flight data, 
and real-time navigation with real-world data. 
Finally, \cref{sec:conclusion} concludes the paper.

\section{Related Work}
\label{sec:related_works}

\myParagraph{Neural Implicit Shape Representations}
Shape representations are central to many problems in computer vision, computer graphics~\cite{Takikawa21cvpr-neuralSdfLod, Tancik20neurips-FFN}, and robotics~\cite{Oleynikova17iros-voxblox, Reijgwart20ral-voxgraph}.
Traditional shape representations such as points clouds, meshes, and voxel-based models, while being well studied and commonly used in robotics, still suffer from several drawbacks.
For example, point clouds lack the ability to encode surface information.
Meshes encode surfaces, but it remains challenging to estimate highly accurate meshes from sensor data collected by a robot~\cite{Rosinol21ijrr-Kimera}. Similarly, the accuracy of voxel-based models is intrinsically limited by the voxel size used for discretization.

Recently, neural implicit shape representations have been developed as effective alternatives to traditional shape representations~\cite{Park19cvpr-deepSDF, Mescheder19cvpr-occupancy-networks,Peng21neurips-shpAsPts,Tancik20neurips-FFN}.
Park\setal~\cite{Park19cvpr-deepSDF} represent shapes as a learned signed distance function using fully-connected neural networks.
Mescheder\setal~\cite{Mescheder19cvpr-occupancy-networks} learn a probability representation of occupancy grids to represent surfaces.
Mildenhall\setal~\cite{Mildenhall20arxiv-nerf} propose \nerf %
and show that by adding view directions as additional inputs,
it is possible to train a network to synthesize novel and photo-realistic views.

Additional studies have investigated the problem of training \nerf with images whose poses are either unknown or known with low accuracy
\cite{Wang21arxiv-nerfmm, Lin21iccv-barf, Meng21iccv-gnerf, Zhang22arxiv-vmrf}.
These methods take several hours or over a day to train and are intended for building a \nerf as opposed to real-time pose estimation with a trained \nerf.
\nerf has also been extended to large-scale \cite{Turki22cvpr-megaNerf, xiangli22eccv-bungeenerf, Tancik22cvpr-blocknerf} and
unbounded scenes \cite{Zhang2020arxiv-nerfpp, Barron22cvpr-mipnerf360},
which has the potential to enable neural representations of large-scale scenes such as the ones 
typically encountered in robotics applications, from drone navigation to self-driving cars.

Slow training and rendering time has been a longstanding challenge for \nerf, with several recent works proposing computational enhancements.
M\"uller\setal~\cite{Mueller22acm-instantngp} use a multi-resolution hash encoding to train a \nerf in seconds and render images on the order of milliseconds.
Additionally, some works have utilized depth information to improve
rendering time~\cite{Neff21cgf-donerf, Chen21iccv-mvsNerf},
and training time~\cite{Kangle21cvpr-dsnerf, Wei21iccv-nerfingmvs, Roessle22cvpr-denseDepthNerf}.
Related to using depth, Clark~\cite{Clark22cvpr-vba} uses a volumetric dynamic B+Tree data structure to achieve real-time scene reconstruction and Yu\setal~\cite{Yu21iccv-plenoctrees} use a scene representation based on octrees.

\myParagraph{Visual Localization}
Classical approaches for visual localization and SLAM in robotics typically use a multi-stage paradigm,
where some sparse representations (such as keypoints) are used to enable tracking and localization~\cite{Klein07ismar-PTAM, Qin19arxiv-VINS-Fusion-odometry,Cadena16tro-SLAMsurvey}.
In some works, instead of sparse keypoints, a dense representation is used to represent the 3D environment~\cite{Newcombe2011iccv-dtam}.
The backend of classical localization methods typically rely on well established estimation-theoretic techniques, such as 
maximum a posteriori (optimization-based) estimation, Kalman filters, particle filters, and grid-based histogram filters;
these techniques enable tracking the pose of the robot over time~\cite{Thrun05book}. 

More recently, localization and mapping have been studied in conjunction with neural implicit representations.
Sucar\setal~\cite{Sucar21iccv-iMAP} propose iMAP, which demonstrates that an MLP can be used to represent the scene 
in simultaneous localization and mapping. %
Zhu\setal~\cite{Zhu22cvpr-niceslam} develop NICE-SLAM, which extends the idea of MLP-based scene representation to larger,
multi-room environments.
Ortiz\setal~\cite{Ortiz22rss-isdf} propose iSDF, which is a continual learning system for real-time signed distance field reconstruction.
iMAP, NICE-SLAM, and iSDF utilize depth information from a stereo camera in addition to color images.
In the RGB-only case, the literature on robot localization based on neural implicit representations is still sparse.
Yen\setal~\cite{Yen20iros-inerf} develop i\nerf, which estimates the pose of a provided image
given a trained \nerf model and an initial pose guess by optimizing a photo-metric loss using back-propagation with respect to the pose; i\nerf requires a good initial guess and the optimization entails a high computational overhead.
Adamkiewicz\setal~\cite{Adamkiewicz21arxiv-nerfNavigation} propose \nerfNav, which uses \nerf to power the  entire autonomy stack of a drone, including estimation, control, and planning. %
Similar to i\nerf, \nerfNav optimizes a loss that includes a photometric loss
along a process loss term induced by the robot dynamics and control actions.
This added process loss enables tracking a path across multiple images, but 
the method still requires a good initial guess and 
incurs a high computation overhead similar to i\nerf.
In this paper, we improve upon i\nerf and \nerfNav and propose \name.
The particle filter backbone~\cite{Dellaert99} of \name allows relaxing the reliance on a good initial estimate to bootstrap localization.

\section{\nerf Preliminaries}
\label{sec:background}

\nerf~\cite{Mildenhall20arxiv-nerf} uses a multilayer perceptron (MLP) to store a radiance field representation of a scene and render novel viewpoints. \nerf is trained on a scene 
given a set of RGB images with known poses and a known camera model. 
At inference time, \nerf renders novel views by predicting the density $\sigma$ and RGB 
color $\vc$ of a point in 3D space given the 3D position and viewing direction of the point. To predict the RGB value of a single pixel, 
\nerf projects a ray $\vr$ from the center point of the camera, through a pixel in the image plane.
Then $n_{\text{coarse}}$ samples are uniformly generated along the ray and $n_{\text{fine}}$ samples are selected based on the estimated $\sigma$ 
of the coarse samples. Volume rendering is then used 
to estimate the color value $\calC(r)$ for the pixel:
\begin{equation}
    \label{eq:volume_rendering}
    \calC(\vr) = \int_{z_{\text{near}}}^{z_{\text{far}}}  T(\vr,z) \sigma(\vr,z) \vc (\vr,z) dz
\end{equation}
where $z_{\text{near}}$ and $z_{\text{far}}$ are bounds on the sampled depth $z$ along the ray $\vr$ and $T(\vr,z)$ is given by: 
\begin{equation}
    \label{eq:T}
    T(\vr,z) = \exp\left(-\int_{z_{near}}^z \sigma(\vr,z^\prime) dz^\prime \right)
\end{equation}
The reader is referred to~\cite{Mildenhall20arxiv-nerf} for a more detailed description. 

\section{\name: Monte Carlo Localization \\ using Neural Radiance Fields}
\label{sec:contribution}

We now present \name, a real-time Monte Carlo localization method that uses \nerf as a map representation.
Given a map $\map$ (encoded by a trained \nerf), RGB input image $\image_t$ at each time $t$, and motion estimates 
$\odom_{t}$ between time $t-1$ and time $t$, \name estimates the 6DoF pose of the robot $\MX_t$ at time $t$. 
In particular, \name  uses a particle filter to estimate the posterior probability $\prob{ \MX_t \mid \map, \image_{1:t}, \odom_{1:t} }$, where $\image_{1:t}$ and $\odom_{1:t}$ are the sets of images and motion measurements collected between the initial time 1 and the current time $t$, respectively. 

Monte Carlo localization~\cite{Dellaert99} relies on a particle filter and
models the posterior distribution $\prob{ \MX_t \mid \map, \image_{1:t}, \odom_{1:t} }$ as a weighted set of $n$ particles:
\begin{equation}
    \label{eq:particles}
    S_t = \left\{ \langle {\MX}\at{i}_t, w\at{i}_t \rangle \mid i=1,...,n \right\}
\end{equation}
where ${\MX}\at{i}_t$ is a 3D pose (represented as a $4\times 4$ transformation matrix in our implementation) associated to the $i$-th particle, and $w\at{i}_t \in [0,1]$ is the corresponding weight.
The particle filter then updates the set of particles at each time instant (as new images and odometry measurements are received) by applying three steps: prediction, update, and resampling. %

\subsection{Prediction Step}

The prediction step predicts the set of particles $S_t$ at time $t$ from the corresponding set of particles $S_{t-1}$ at time $t-1$, given a measurement $\odom_{t}$ of  the robot motion between time 
$t-1$ and time $t$; the measurement is typically provided by some odometry source (\eg wheel or visual odometry) or obtained by integrating the robot dynamics; in our implementation, we either use visual-inertial odometry or integrate the robot dynamics, depending on the experiment.
When a measurement of the robot's relative motion $\odom_{t}$ is received, the set of particles can be updated by sampling new particles using the motion model $\prob{ \MX_t \mid \MX_{t-1}, \odom_{t} }$. 
While the particle filter can accommodate arbitrary motion models, here we adopt   
a simple model 
that updates the pose of each particle according to the motion $\odom_{t}$ and then adds Gaussian noise to account for odometry errors:
\begin{equation}
    \label{eq:predict}
    \MX_{t} = \MX_{t-1} \cdot \odom_{t} \cdot \MX_\epsilon 
    \quad,
    \quad
    \MX_\epsilon = \expmap{\vdelta},
\end{equation}
where $\MX_\epsilon$ is the prediction noise, $\expmap{\cdot}$ is the exponential map for $\SEthree$ (the Special Euclidean group), and $\vdelta \in \Real{6}$ is a normally distributed vector with zero mean and covariance 
$\diag{ \sigma_R^2 \cdot \eye_3, \sigma_t^2 \cdot \eye_3 }$, where $\sigma_R$ and $\sigma_t$ are the rotation and translation noise standard deviations, respectively. 

\subsection{Update Step}

The update step uses the camera image $\image_t$ collected at time $t$ to update the particle weights $w\at{i}_t$.
According to standard Monte Carlo localization~\cite{Dellaert99}, we update the weights using the measurement likelihood 
$\prob{\image_t \mid \MX\at{i}_t, \map}$, which models the likelihood of taking an image $\image_t$ from pose $\MX\at{i}_t$ in the map $\map$. 
We use a heuristic function to approximate the measurement likelihood as follows:

\beq
    \label{eq:weight}
    w\at{i}_t =
     \left(\frac{M}
     {\sum_{j=1}^M (\image_t(\vp_j) - C(\vr(\vp_j,\MX\at{i}_t)))^2}
     \right)^4
\eeq
where $\vr(\vp_j,\MX\at{i}_t)$ computes the ray emanating from pixel $\vp_j$ when the robot is at pose $\MX\at{i}_t$, 
and $\image_t(\vp_j)$ is the image intensity at pixel $\vp_j$.
Intuitively, eq.~\eqref{eq:weight} compares the collected image $\image_t$ with the image $\calC(\vr)$ predicted by the \nerf map and assigns low weights to particles where the two images do not match.   
For efficient computation, we 
compute the weight update~\eqref{eq:weight} only using a subset of $M$ pixels randomly sampled from $\image_t$.
Weights are then normalized to sum up to 1. 

\subsection{Resampling Step}

After the update step, we resample $n$ particles from the set $S_t$ with replacement, where each particle is sampled with probability $w\at{i}_t$. As prescribed by standard particle filtering, 
the resampling step allows retaining 
 particles that are more likely to correspond to good pose estimates while discarding less likely hypotheses. 

\subsection{Computational Enhancements and Pose Estimate}

\myParagraph{Particle Annealing}
To improve convergence of the filter and reduce the  computational load, we automatically adjust the prediction noise $(\sigma_R,\sigma_t)$ and the number of particles $n$ over time.
As shown in~\cref{sec:experiments}, this leads to computational and accuracy improvements.
The prediction noise and number of particles are updated as shown in \cref{alg:refine}. 
In particular, we use the standard 
deviation of the particles' position 
$\sigma_{S_t}$ to characterize the spread of the particles in the filter at time $t$ and reduce the prediction noise and the number of particles (initially set to $\sigmaRinit$, $\sigmatinit$, and $n_{\text{\footnotesize init}}$) when the spread falls below given thresholds ($\alpharef$ and $\alphasuperref$ in~\cref{alg:refine}).

\begin{algorithm}
    \caption{Particle Annealing \label{alg:refine}}
    \textbf{Input:} $\sigmaRinit, \;\; \sigmatinit, \;\; \sigma_{S_t}, \;\; n_{\text{\footnotesize init}}$
    \begin{algorithmic}
    \State $\sigma_R \gets \sigmaRinit$
    \State $\sigma_t \gets \sigmatinit$
    \State $n \gets n_{\text{\footnotesize init}}$
    \If{$\sigma_{S_t} < \alphasuperref$}
        \State $\sigma_R \gets \frac{\sigmaRinit}{4}$ 
        ,\quad $\sigma_t \gets \frac{\sigmatinit}{4}$
        ,\quad 
        $n \gets \nreduced$
    \ElsIf{$\sigma_{S_t} < \alpharef$}
        \State $\sigma_R \gets \frac{\sigmaRinit}{2}$
        ,\quad
         $\sigma_t \gets \frac{\sigmatinit}{2}$
         ,\quad
        $n \gets \nreduced$
    \Else
        \State $\sigma_R \gets \sigmaRinit$,\quad $\sigma_t \gets \sigmatinit$
    \EndIf
    \end{algorithmic}
    \end{algorithm}

\myParagraph{Obtaining a Pose Estimate from the Particles}
Besides computing the set of particles, \name returns a single pose estimate $\hat{\MX}_t$ that is computed as a weighted average of the particle poses. In particular, the position portion of $\hat{\MX}_t$ is simply the weighted average of the positions of the particles in  $S_t$. 
The rotation portion of $\hat{\MX}_t$ is found by solving the geodesic $L_2$ single rotation averaging problem.
The reader is referred to \cite{Hartley13ijcv} and \cite{Manton04icarcv-centreMassLie} for details on rotation averaging.

\section{Experiments}
\label{sec:experiments}

We evaluate \name on three sets of experiments: 
(i) pose estimation from a single image using the LLFF dataset~\cite{Mildenhall19tog-llff} 
 given either a poor initial guess or no initial 
guess, where we benchmark against iNeRF~\cite{Yen20iros-inerf} (\cref{sec:experiments_inerf}), 
(ii) pose estimation over time using synthetic data from Blender~\cite{Blender18-blender}, where
we  benchmark against \nerfNav~\cite{Adamkiewicz21arxiv-nerfNavigation} (\cref{sec:experiments_bohg}), 
and (iii) a full system demonstration where we perform real-time pose tracking using data collected by a Clearpath Jackal UGV (\cref{sec:experiments_a1}).

\subsection{\mbox{Single-image Pose Estimation: Comparison with iNeRF}}
\label{sec:experiments_inerf}

\myParagraph{Setup}
To show \name's ability to quickly localize given a camera image and from a poor initial guess, we use the same evaluation protocol used in iNeRF~\cite{Yen20iros-inerf}.
Using 4 scenes (Fern, Fortress, Horns, and Room) from the LLFF dataset \cite{Mildenhall19tog-llff}, we pick 5 random images 
from each dataset and estimate the pose of each image. 
For this experiment, both \name and iNeRF use the same pre-trained weights from NeRF-Pytorch \cite{Lin20-nerfpytorch}.
As in~\cite{Yen20iros-inerf}, we give iNeRF an initial pose guess $\MX_{\small\text{iNeRF}}$.
The rotation component of $\MX_{\small\text{iNeRF}}$ is obtained by randomly sampling 
an axis from the unit sphere and rotating about that axis by a uniformly sampled angle between [-40$^{\circ}$, 40$^{\circ}$] with respect to the ground truth rotation. 
The position portion  of $\MX_{\small\text{iNeRF}}$ is obtained by uniformly perturbing the ground truth position along each axis by a random amount between [-0.1 m, 0.1 m]. 
We set iNeRF to use 2048 interest region points ($M$ = 2048) as suggested in \cite{Yen20iros-inerf}. 
Interest regions are found using keypoint detectors and sampling from a dilated mask around those keypoint. 

Since \name uses a distribution of particles, we uniformly 
distribute the initial particles' poses using: %

\begin{equation}
    \label{eq:inital_pose}
    \MX\at{i}_0 = \MX_{\small\text{iNeRF}} \cdot \expmap{\vdelta}
\end{equation}
where the entries corresponding to the rotation component and the translation component of $\vdelta$ 
are sampled from a uniform distribution in the range [-40$^{\circ}$, 40$^{\circ}$] and [-0.1 m, 0.1 m], respectively. 
Since we only test on a static image, we set the motion model of \name to be a zero-mean Gaussian distribution whose standard deviation decreases according to~\cref{alg:refine}.
\name is initialized with 300 particles which reduces to 100 during \refining. 
We set \name to use 64 ($M$ = 64) randomly sampled image pixels per particle. 

\myParagraph{Results}
We plot the fraction of estimated poses with position and rotation error less than 5 cm and 5$^{\circ}$ in \cref{fig:inerf_ratio_rot} 
and \cref{fig:inerf_ratio_position}, respectively.
Since the computational cost of an iNeRF iteration is different from an iteration of \name
(due to number of particles and different values of $M$) 
we plot performance against the number of \nerf forward passes. 
\name achieves higher accuracy than iNeRF in terms of both position and rotation. 

We also plot the average rotation error and average position error for all 20 trials in \cref{fig:rot_error_inerf} 
and \cref{fig:position_error_iner} respectively. In our experiments, the position estimate from iNeRF would occasionally diverge or 
reach a local minimum and thus the average position error for iNeRF actually increases over time. 
On a laptop with an RTX 5000 GPU, 
the update step for \name runs at 0.6 Hz for 300 particles which then accelerates to 1.8 Hz during \refining when the number of particles drops to 100. 
\name runs approximately 55 seconds per trial. 
As an ablation study of our \refining process (\cref{alg:refine}), 
we also include results of \name without \refining. Using \refining shows the most benefit for position accuracy and allows update steps 
to occur at a faster rate due to the decreased number of particles.

\newcommand{\myheight}{4.9cm}
\begin{figure*}
    \begin{subfigure}[b]{1\columnwidth}
        \includegraphics[height=\myheight]{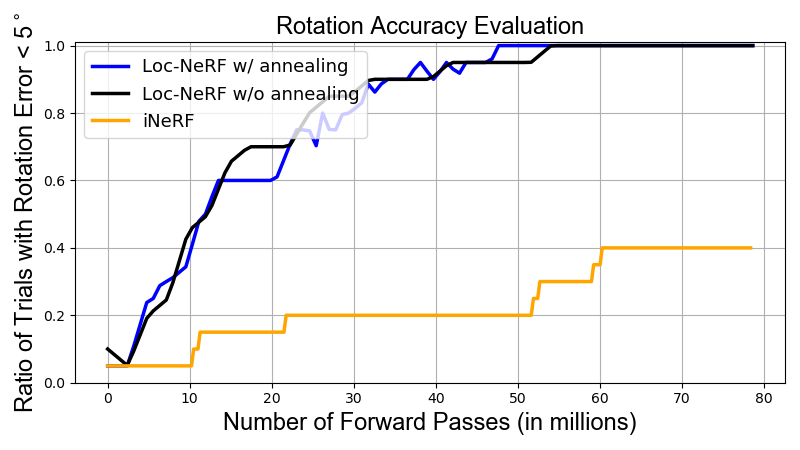}
        \caption{ }
        \label{fig:inerf_ratio_rot}
    \end{subfigure}
    \hfill
    \begin{subfigure}[b]{1\columnwidth}
        \includegraphics[height=\myheight]{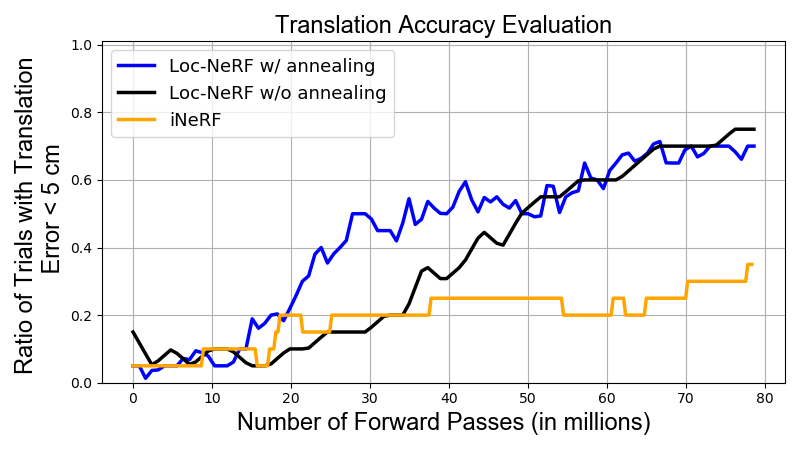}
        \caption{ }
        \label{fig:inerf_ratio_position}
    \end{subfigure}
    \begin{subfigure}[b]{1\columnwidth}
        \includegraphics[height=\myheight]{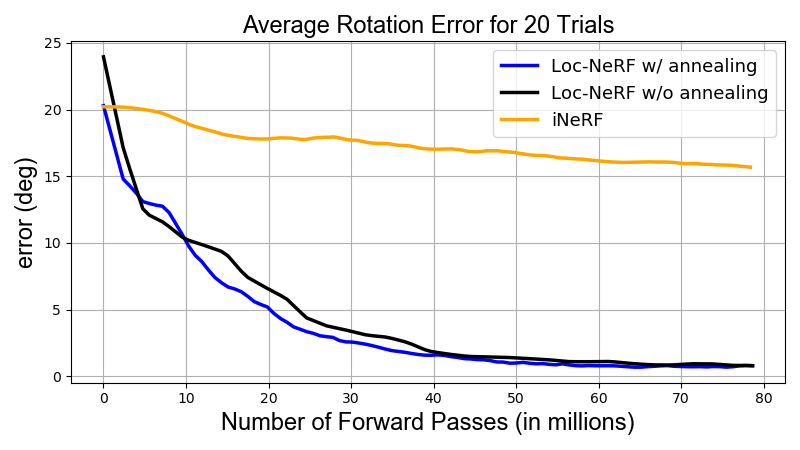}
        \caption{ }
        \label{fig:rot_error_inerf}
    \end{subfigure}
    \hfill
    \begin{subfigure}[b]{1\columnwidth}
        \includegraphics[height=\myheight]{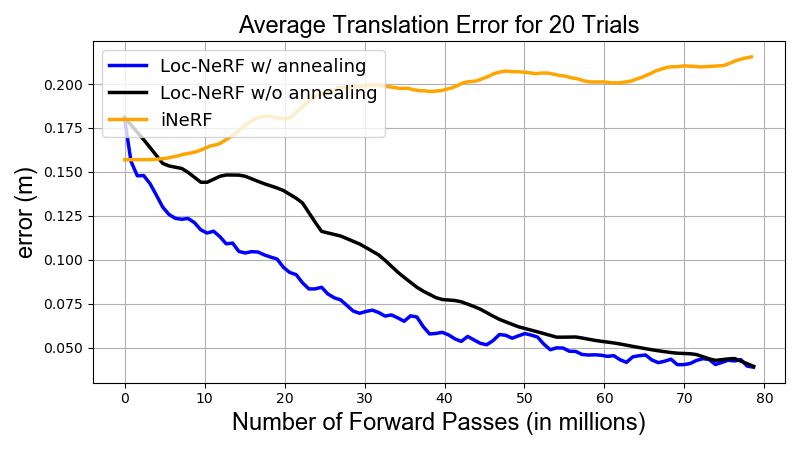}
        \caption{ }
        \label{fig:position_error_iner}
    \end{subfigure}
    \caption{Evaluation of \name and iNeRF on 20 camera poses from the LLFF dataset. As an ablation study 
    of our \refining step, we also include results of \name without using~\cref{alg:refine}. 
    (a) Ratio of trials with rotation error $<$ 5$^{\circ}$. (b) Ratio of trials with translation error $<$ 5 cm.
    (c) Average rotation error. (d) Average translation error.}
\end{figure*}

\begin{figure*}
    \begin{subfigure}[b]{1\columnwidth}
        \includegraphics[height=\myheight]{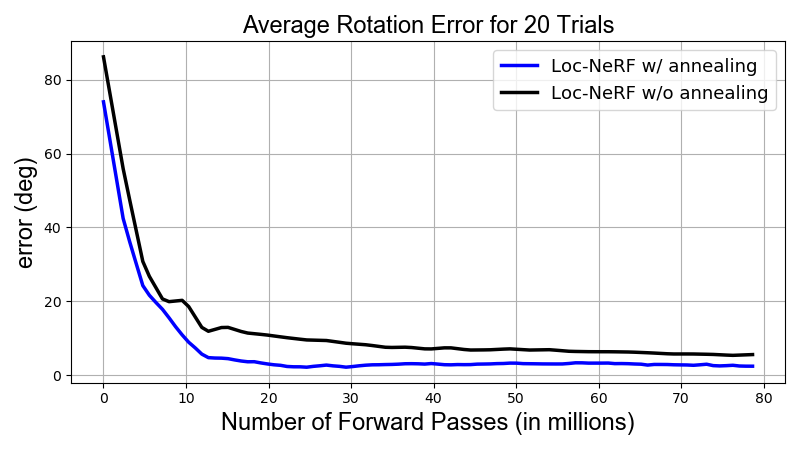}
        \caption{ }
        \label{fig:global_rot}
    \end{subfigure}
    \hfill
    \begin{subfigure}[b]{1\columnwidth}
        \includegraphics[height=\myheight]{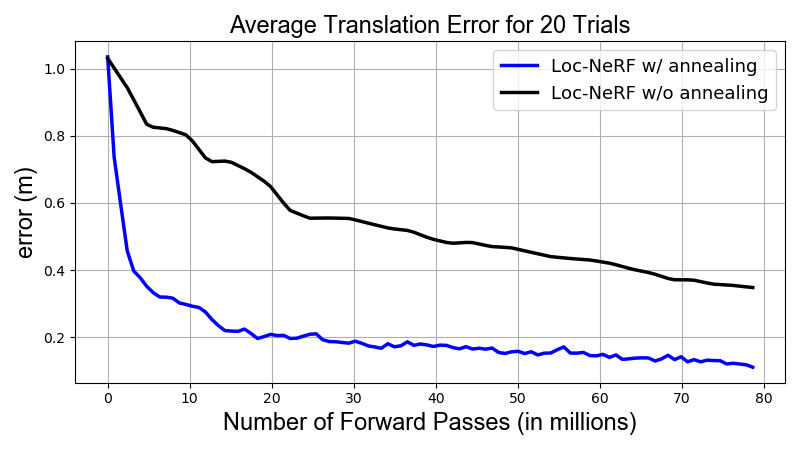}
        \caption{ }
        \label{fig:global_position}
    \end{subfigure}
\caption{Evaluation of \name on 20 camera poses from the LLFF dataset without an initial guess for the unknown pose. 
(a) Average rotation error. (b) Average translation error.}
\end{figure*}

We also demonstrate for the first time that global localization can be performed with NeRF. We repeat a similar experiment as before with 
LLFF data except now we generate an offset translation by translating the ground truth position along each axis 
by a random amount between [-1 m, 1 m] and generate 
a random  distribution of particles in a $2\times 2 \times 2$~m cube about that offset. We then sample the yaw angle from a uniform distribution in $[-180^{\circ},+180^{\circ}]$, while we initialize the roll and pitch to the ground truth; 
the latter is done to mimic the setup where we localize using visual-inertial sensors, in which case the IMU makes roll and pitch directly observable.
Note that \name still optimizes the particles in a full 6DoF state. 
We increase the initial number of particles to 600 which 
drops to 100 during \refining and reduce $M$ to 32. Results of average rotation and translation error from 20 trials 
are provided in \cref{fig:global_rot} and \cref{fig:global_position}. \name is able to converge to an accurate pose estimate while performing global localization. 
The \refining process is shown to enable significant improvement for position accuracy and also improves rotation accuracy.
iNeRF is unable to produce a valid result for global localization and is thus not included in the figure. 

\subsection{Pose Tracking: Comparison with \nerfNav}
\label{sec:experiments_bohg}

\begin{figure}
\centering
    \includegraphics[width=0.4\textwidth]{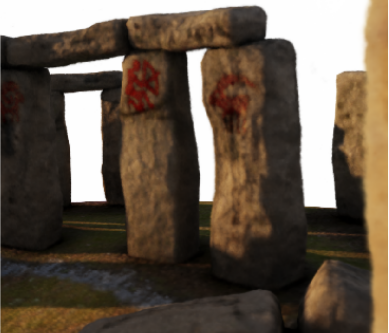}
    \caption{Example of NeRF rendering of a scene from Stonehenge. \label{fig:stonehindge}}
\end{figure}

\myParagraph{Setup}
\nerfNav \cite{Adamkiewicz21arxiv-nerfNavigation} performs localization using simulated image streams of Stonehenge
recreated in Blender, as if they were collected by a drone flying across the scene
(\cref{fig:stonehindge}). 
For this experiment, both \name and \nerfNav use the same pre-trained weights from torch-ngp \cite{Tang22-torch-ngp}. 
We use the same trajectory and sensor images for evaluating \name and \nerfNav. 
The prediction step for \name uses the same dynamical model estimate 
of the vehicle's motion that \nerfNav uses for their process loss.  For each image, we run \name for the equivalent number of forward passes as \nerfNav. 
We run \nerfNav for 300 iterations per image with $M = 1024$. 
We use 200 particles for \name with $M = 64$  and run 24 update steps per image. 

\begin{figure}
\centering
    \includegraphics[width=0.49\textwidth]{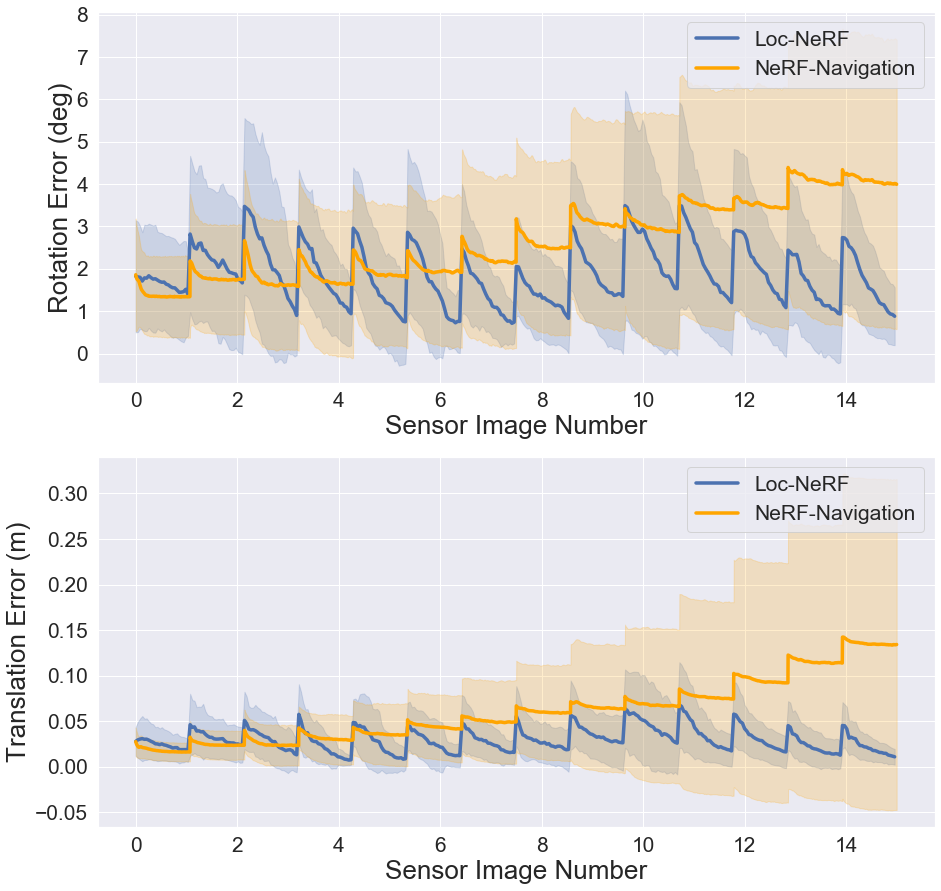}
    \caption{Translation and rotation error of Loc-NeRF and \nerfNav averaged 
    over 18 trials. The shaded area shows one standard deviation above and below the mean error. 
    The area between each sensor image number shows the optimization steps. Spikes at the beginning of each sawtooth show error when an image is first recieved 
    and the pose is forward propaged with a dynamics model, and the bottom of each sawtooth represents the final 
    pose estimate after optimization. For a fair 
    comparison, both methods run the same number of forward passes for each camera image. \label{fig:nerfnav_results}}
\end{figure}

\myParagraph{Results}
\cref{fig:nerfnav_results} shows position and rotation error respectively 
for a simulated drone course over 18 trials. 
Note that since \nerfNav uses a similar photometric loss as iNeRF ---which requires a good initial guess--- we assume the starting pose 
of the drone is well known even though that is not a requirement for \name. The process loss of \nerfNav gives it added robustness to portions of the trajectory 
where the NeRF rendering is of lower quality. 
However, \name is still able to achieve lower errors for both position and rotation on average and is able to recover from inaccurate pose estimates. 

\subsection{Full System Demonstration}
\label{sec:experiments_a1}
Finally, we demonstrate our full system running in real-time on real data collected by a robot. We pre-train a NeRF model using NeRF-Pytorch \cite{Lin20-nerfpytorch}
with metric scaled poses and images from a Realsense d455 camera carried by a person. 
To run \name, we use a Realsense d455 as the vision sensor mounted on a Clearpath Jackal UGV. The prediction step for \name is performed using VINS-Fusion~\cite{Qin19arxiv-VINS-Fusion-odometry}. We log images and IMU data from the Jackal and then run VINS-Fusion and \name 
simultaneously on a laptop with an 
RTX 5000 GPU.

We initialize particles across a $1\times 0.5\times3.5$ m area with a uniformly distributed yaw in [-180$^\circ$,+180$^\circ$] 
and uniformly distributed roll and pitch in [-2.5$^\circ$,+2.5$^\circ$] (again, the latter are directly observable from the IMU).
The prediction step runs at the nominal VIO rate of 15 Hz. \name starts with 
400 particles which reduces to 150 during particle annealing. We set $M$ to 32. With 400 particles the update step runs at approximately 0.9 Hz and then 
accelerates to 2.5 Hz with 150 particles during \refining.  
In this experiment, the particles quickly converge enough to trigger the \refining stage 
after about 6 update steps.

\begin{figure}
    \includegraphics[width=0.48\textwidth]{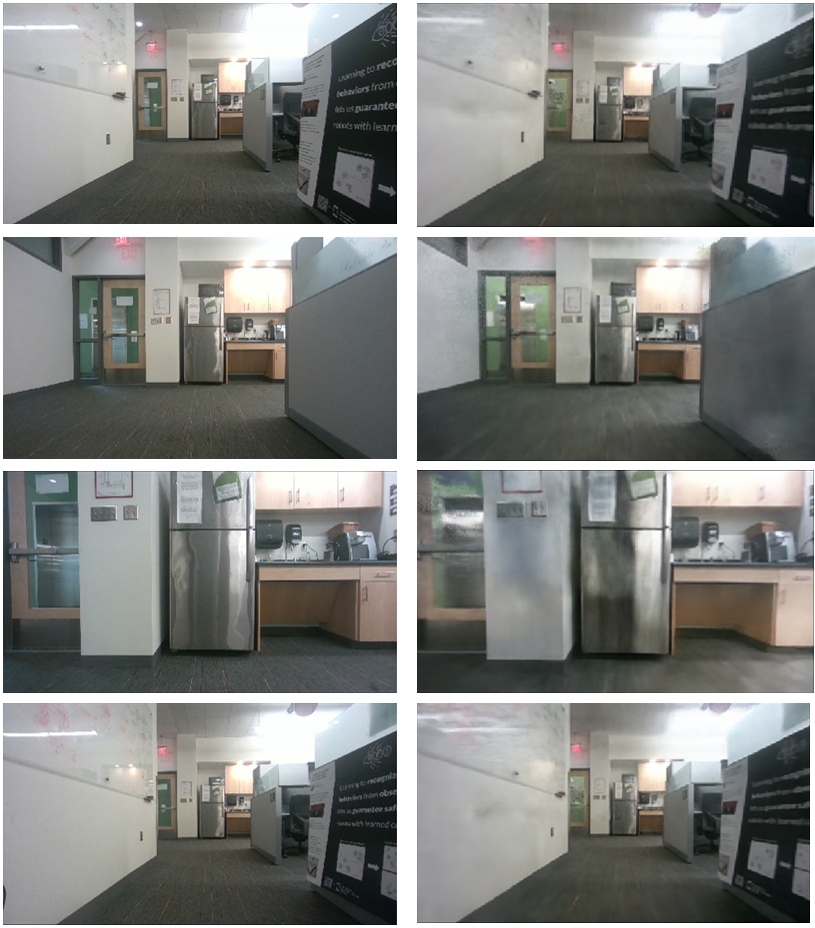}
    \caption{Left Column: true images viewed by the camera. Right Column: NeRF-rendered images using the pose estimate from \name. 
    Images correspond to update steps number 20, 40, 60, and 100 which occur at 13, 20, 28, and 44 seconds into the experiment, respectively.}
    \label{fig:nerf_compare}
\end{figure}

To qualitatively demonstrate that \name converges to the correct pose, we render a full image from \nerf using the pose estimated by \name and compare it with the corresponding camera image. 
We provide results from this test in~\cref{fig:nerf_compare} at selected points in the trajectory. 

\section{Conclusion}
\label{sec:conclusion}

We propose \name, a Monte Carlo localization approach that uses a Neural Radiance field (\nerf) as a map representation.
We show how to incorporate \nerf in the update step of the filter, while the prediction step can be done 
using existing techniques (\eg visual-inertial navigation or by leveraging the robot dynamics).
We show \name is the first approach to perform localization with \nerf from a poor initial guess, and can be used for global localization. 
We have also demonstrated the ability to perform real-time localization with \name on a real-world robotic platform. 
Future work includes using adaptive techniques to adjust the number of particles~\cite{Fox99aaai}
as well as scaling up localization to larger environments using bigger NeRF models 
such as~\cite{Turki22cvpr-megaNerf} and~\cite{Tancik22cvpr-blocknerf}. Additionally, computation time can be 
reduced by leveraging recent work in faster NeRF rendering such as~\cite{Tang22-torch-ngp}.

\section*{Acknowledgement}
The authors would like to gratefully acknowledge Timothy Chen, Michal Adamkiewicz, and all the authors of NeRF-Navigation who assisted us with benchmarking their work. 
We also acknowledge Jared Strader for assisting with collecting experimental data.

\bibliographystyle{IEEEtran}

\end{document}